\documentclass[]{interact}
\usepackage{xcolor}
\usepackage{mathtools}
\usepackage{epstopdf}
\usepackage[caption=false]{subfig}

\usepackage[natbibapa,nodoi]{apacite}
\theoremstyle{plain}

\theoremstyle{definition}

\theoremstyle{remark}

\begin{document}

\articletype{RESEARCH ARTICLE}

\title{Automated Crack Propagation Measurement On Asphalt Concrete Specimens Using an Optical Flow-Based Deep Neural Network}

\author{
\name{Zehui Zhu\textsuperscript{a}\thanks{CONTACT Zehui Zhu. Email: zehuiz2@illinois.edu} and Imad L. Al-Qadi\textsuperscript{a}}
\affil{\textsuperscript{a} Illinois Center for Transportation, University of Illinois Urbana-Champaign, Rantoul, IL 61866 USA}
}

\maketitle

\begin{abstract}
This article proposes a deep neural network, namely CrackPropNet, to measure crack propagation on asphalt concrete (AC) specimens. It offers an accurate, flexible, efficient, and low-cost solution for crack propagation measurement using images collected during cracking tests. CrackPropNet significantly differs from traditional deep learning networks, as it involves learning to locate displacement field discontinuities by matching features at various locations in the reference and deformed images. An image library representing the diversified cracking behavior of AC was developed for supervised training. CrackPropNet achieved an optimal dataset scale F-1 of 0.755 and optimal image scale F-1 of 0.781 on the testing dataset at a running speed of 26 frame-per-second. Experiments demonstrated that low to medium-level Gaussian noises had a limited impact on the measurement accuracy of CrackPropNet. Moreover, the model showed promising generalization on fundamentally different images. As a crack measurement technique, the CrackPropNet can detect complex crack patterns accurately and efficiently in AC cracking tests. It can be applied to  characterize the cracking phenomenon, evaluate AC cracking potential, validate test protocols, and verify theoretical models.
\end{abstract}


\section{Introduction}
Approximately 95 percent of paved roads in the United States are surfaced with asphalt. Cracking is a common mode of failure in asphalt concrete (AC) pavements. Many tests have been developed to assess the cracking potential of AC materials. Accurate monitoring of crack initiation and propagation during testing is crucial.

\begin{figure}[!t]
    \centering
    \includegraphics[width=2.5in]{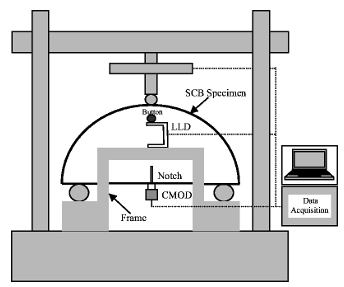}
    \caption{Low-temperature SCB test setup \citep{li2010using}.}
    \label{fig_scb}
\end{figure}

Contact tools like a linear variable differential transformer (LVDT), extensometers, strain gauges, and crack mouth opening displacement (CMOD) clip gauges are the most widely used methods to monitor crack propagation and opening in AC cracking tests. However, these tools only provide localized information, as the measurement location must be decided before testing. As shown in Figure \ref{fig_scb}, the CMOD clip gauge is attached at the bottom of the specimen when conducting the low-temperature semi-circular bending (SCB) test \citep{li2010using}. As such, the crack opening is only recorded at that location. Similarly, the above-mentioned contact techniques may only provide indirect crack propagation and opening measurements. For example, the load-line displacement (LLD), measured by an extensometer mounted vertically at the surface of the specimen, is used to estimate crack propagation speed. However, a such approximation is insufficient to describe the cracking phenomenon, as a crack tends to choose a path around the aggregate as it grows \citep{doll2017damage}. In addition, orienting the contact devices is time-consuming and requires experience, especially on small specimens. Moreover, routine calibrations are needed to ensure accurate measurement. Therefore, developing an easy-to-use, accurate, and full-field crack measurement technique for AC cracking tests is imperative. 

Low-level computer vision-based crack detection methods have been proposed. The most popular algorithms include thresholding, image segmentation, filtering, and blob extraction \citep{hartman2004evaluating,oliveira2009automatic,ying2010beamlet,zhang2013matched,wang2018pavement}. However, the limitation of these methods is obtaining accurate results under complex imaging environments. Deep learning, especially deep convolution neural network (CNN), has been widely used to detect and categorize cracks \citep{cha2017deep, zhang2016road, zhang2017automated, fei2019pixel}. However, these models were mainly developed for visible mature cracks; the ground truth verification relied on visual recognition. This renders the models mentioned above unsuitable for monitoring crack propagation in AC cracking tests, where small cracks in the early stages are critical but often difficult to visualize.

The digital image correlation (DIC) technique has the potential to overcome these challenges. The DIC is an optical method that measures full-field displacement and strain. Because surface cracks are defined as displacement field discontinuities, cracks with varying sizes could be located, given an accurate displacement field. A few attempts have been made to measure cracks using DIC. Due to complex crack growth, locating cracks based on DIC-measured displacement or strain field is a challenge. Current methods rely on the strain or displacement thresholding, which requires significant post-processing efforts and empirical knowledge \citep{buttlar2014digital,safavizadeh2017dic}. In addition, DIC analysis involves computationally expensive optimization, making it unsuitable for real-time applications such as crack propagation measurement in AC testing, where hundreds or thousands of images need to be analyzed. These limitations have limited the implementation of DIC as an automated crack measurement technique for AC cracking tests.

This article proposes a deep neural network to automatically measure crack propagation during testing based on the optical flow concept. Compared to the existing techniques discussed above, it offers an accurate, flexible, efficient, and low-cost solution. It can accurately measure crack propagation from hundreds of images collected by low-cost cameras in less than one minute.

This paper has seven sections, and they are organized as the following: section one discusses the background and motivation of this study; section two presents the development of the database; section three introduces the architecture of the deep neural network; section four explains the training strategy; section five presents the evaluation of the proposed network; section six discusses advantages and possible applications of the CrackPropNet, and the conclusions and recommendations are presented in section seven.

\section{Data Preparation}

\begin{figure}[!t]
    \centering
    \includegraphics[width=5.5in]{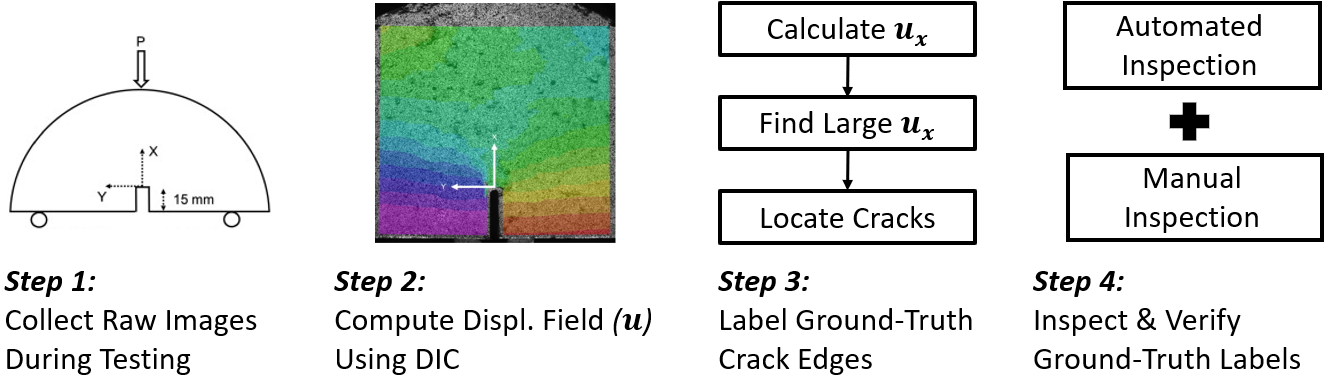}
    \caption{Data preparation procedure.}
    \label{fig_data}
\end{figure}

As shown in Figure \ref{fig_data}, the data preparation process consists of four steps:
\begin{enumerate}
    \item Collect raw images;
    \item Compute displacement fields using DIC;
    \item Label ground-truth crack edges;
    \item Inspect and verify ground-truth labels.
\end{enumerate}

Implementation details are described in the following sections.

\subsection{Raw Image Collection}
\begin{figure}[!t]
    \centering
    \includegraphics[width=6in]{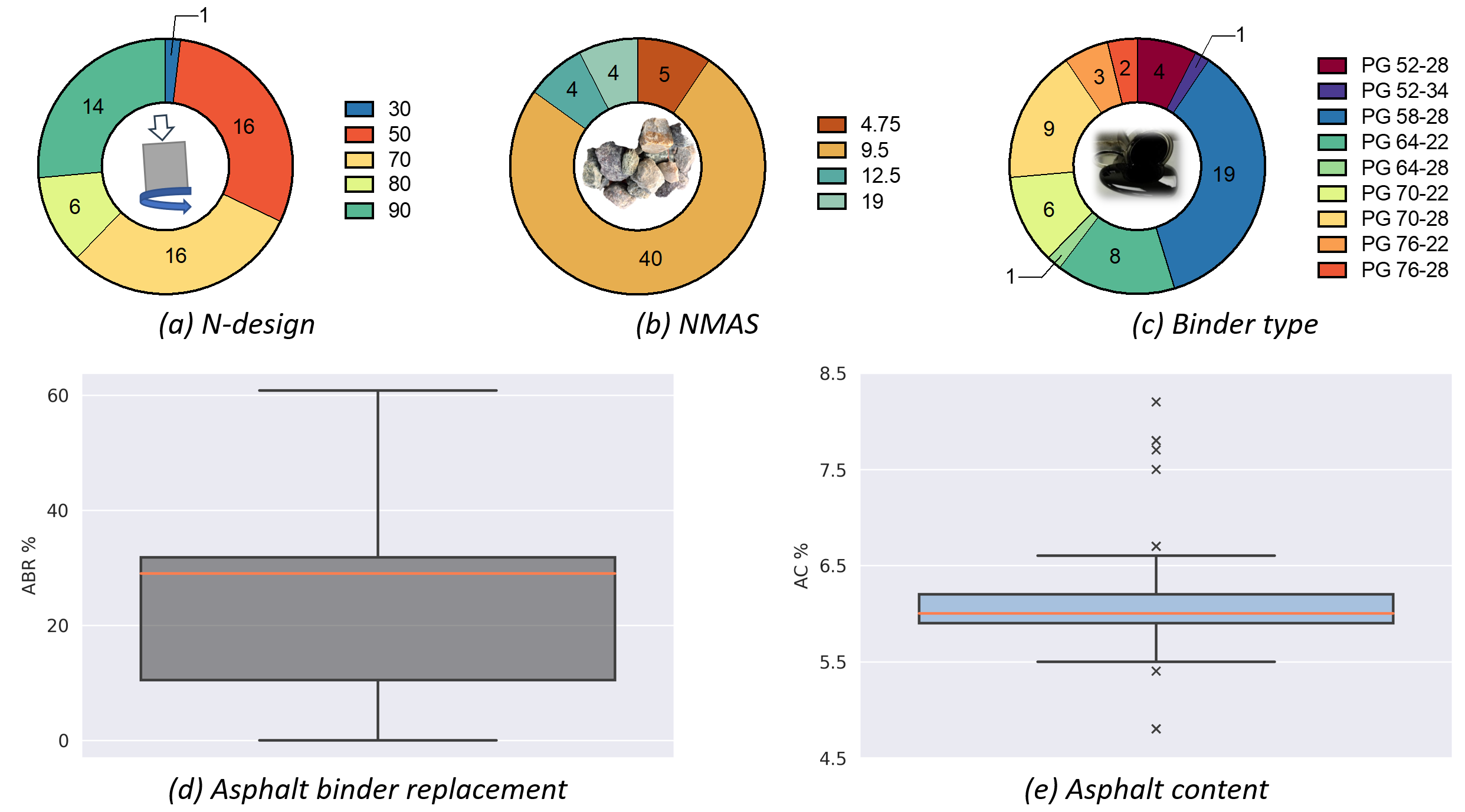}
    \caption{Mix properties.}
    \label{fig_db}
\end{figure}

\begin{figure}[!t]
    \centering
    \includegraphics[width=2.5in]{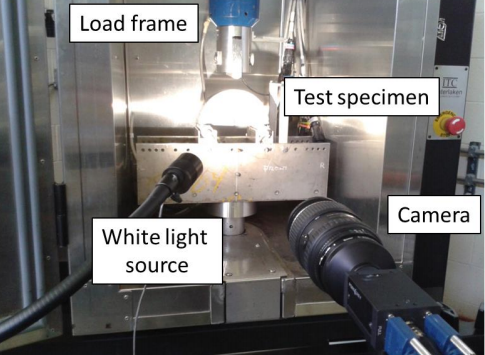}
    \caption{Experiment set up.}
    \label{fig_dic_setup}
\end{figure}

Raw images were collected while conducting the Illinois Flexibility Index Test (I-FIT), as shown in Figure \ref{fig_dic_setup} \citep{ozer2016development, ozer2016fracture}. An extensive testing program covering a wide range of testing conditions and materials was developed. The goal was to develop an extensive image database covering AC’s diversified cracking behavior. All experiments were displacement controlled. Load-line displacement was used on room-temperature tests, while CMOD was used under low temperatures to provide better crack propagation stability \citep{doll2017damage}.  
Two different testing temperatures: -12 and 25\textdegree{}C; and four different loading rates: 0.7, 6.25, 25, and 50 mm/min were considered. The fracture behavior of AC is time- and temperature-dependent. A more brittle failure is expected at lower temperatures or higher loading rates \citep{al2015testing}.
As shown in Figure \ref{fig_db}, a total of 53 AC mixes were tested. They had different N-designs, binder types and content, aggregate mineralogy, and the amount of recycled materials. The I-FIT specimens were prepared either from lab-compacted cylindrical pills or field cores. All specimens had SCB geometry, as shown in Figure \ref{fig_scb}, while their thickness, notch length, and air void range from 25mm to 60mm, 10mm to 35mm, and 1\% to 12\%, respectively. For each I-FIT specimen, a speckle pattern consisting of a white layer of paint and a random black pattern on top was applied \citep{doll2017damage}. 

Two CCD (Charge Coupled Device) cameras were positioned perpendicularly to the surface of the I-FIT specimen to collect images during the test: a Point Grey Gazelle 4.1MP Mono ($2048 \times 2048$ pixels, 150 frames per second-fps) and an Allied Vision Prosilica GX6600 ($6576 \times 4384$ pixels, 4 fps) with a Tokina AT-X Pro Macro 100 2.8D lens. The Gazelle has a faster acquisition rate but a lower resolution than the Prosilica. The former is generally used in experiments where the materials can be considered homogeneous, while the latter aims to study damage zone evolution in heterogeneous materials such as AC. The database intentionally includes images taken from cameras with significantly different resolutions to ensure better generalization of the deep neural network. 

\subsection{Compute Displacement Fields Using DIC}

The displacement fields were first computed using DIC. The DIC works by tracking pixels in a sequence of images. This  is  achieved  using  area-based  matching,  which  extracts  gray  value  correspondences based on their similarities. First, a reference image was taken at the unloaded state, and an area of interest was selected. Then, a subset of pixels is compared to a deformed image taken at a loaded state to identify the best match. Finally, the deformation of a point in the subset can be computed using Equation \ref{eqn_shape_fn}, which allows for translation, rotation, shear, and combinations. 

\begin{figure}[!t]
    \centering
    \includegraphics[width=4.in]{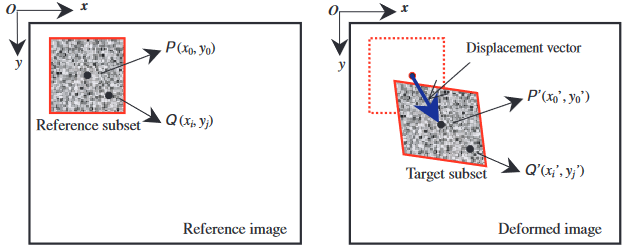}
    \caption{Area-based matching \citep{pan2009two}.}
    \label{fig_match}
\end{figure}

\begin{equation}
    \begin{cases}
    x_i' = x_i+u+u_x \Delta x +u_y \Delta y\\
    y_j' = y_j+v+u_x \Delta x +u_y \Delta y
\end{cases}
\label{eqn_shape_fn}
\end{equation}

As shown in Figure \ref{fig_match}, $x_i$ and $y_j$ are Cartesian coordinates of a point $Q(x_i, y_j)$ in the reference image; $x_i'$ and $y_j'$ refer to its coordinates in the deformed image; $u$ and $v$ denote the corresponding displacement components of the reference subset center $P(x_0,y_0)$ in the x- and y- direction, respectively; $u_x$, $u_y$, $v_x$, $v_y$ are the first-order displacement gradients of the reference subset; $\Delta x = x_i - x_0$ and $\Delta y = y_j - y_0$. To provide adequate spatial resolution to resolve the displacement distribution between and within aggregate particles, the subset size used for correlation was carefully chosen for each test following the algorithm proposed by \cite{pan2008study}.

\subsection{Ground-Truth Crack Edges Labeling}

\begin{figure}[!t]
    \centering
    \includegraphics[width=3.4in]{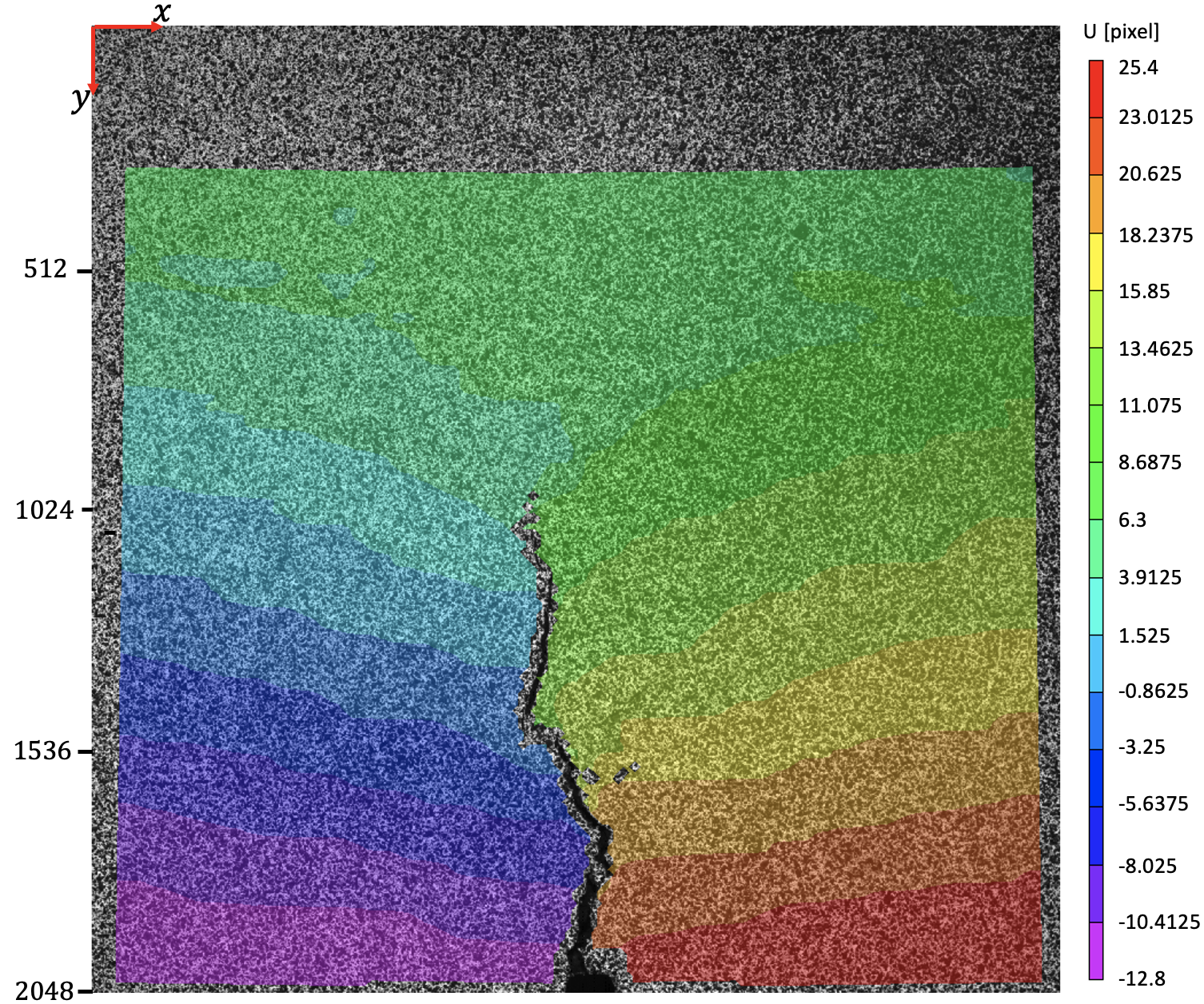}
    \caption{Contour plot of horizontal displacement ($u$).}
    \label{fig_disp_field}
\end{figure}

Once the displacement field was obtained, potential crack edges could be located following the method proposed by \cite{zhu2022}. Figure \ref{fig_disp_field} shows the displacement field ($u$) contour plot measured by DIC for an I-FIT specimen surface, where a crack is visible in the area of interest. The reference and deformed images were taken with the Gazelle camera when conducting the I-FIT test at 25\textdegree{}C with a 50mm/min loading rate. A subset size of $23 \times 23$ with a correlation point spacing of 11 pixels was used for DIC analysis. This resulted in a spatial resolution of approximately 25 $\mu\textnormal{m/pixel}$ and produced roughly $175 \times 157$ square microns in the area of interest. 

\begin{figure}[!t]
    \centering
    \includegraphics[width=2.8in]{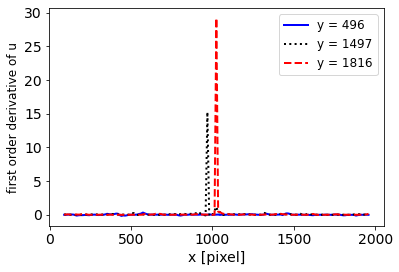}
    \caption{First-order derivative of the opening displacement field $u_x$ along three discrete $y$ axes.}
    \label{fig_disp_first_order}
\end{figure}

\begin{figure}[!t]
    \centering
    \includegraphics[width=2.9in]{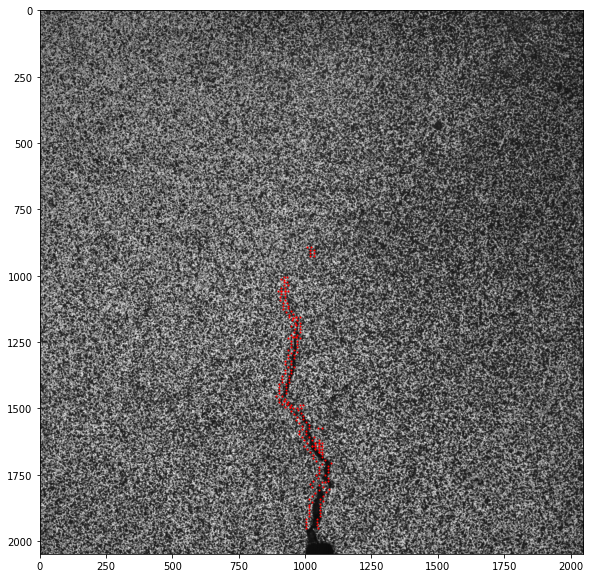}
    \caption{Deformed image with crack edges marked in red dots.}
    \label{fig_crack}
\end{figure}

Then, the first-order derivative of the opening displacement $u_x$ was obtained by filtering the displacement field with a $[-1,1]$ kernel. For example, Figure \ref{fig_disp_first_order} plots $u_x$ along three discrete $y$ axes ($y=$ $496$, $1497$, and $1816$). A large $u_x$ indicates the material separation between the two correlation points, suggesting a crack may present. Thus, potential crack edges could be found by locating the corresponding correlation points in the deformed image, as shown in Figure \ref{fig_crack}. Please note that the marked crack edges may not match the actual crack edges exactly because the correlation point spacing is usually larger than 1 pixel in DIC analysis. However, the effect is negligible as the spatial resolution is typically smaller than 25 $\mu\textnormal{m/pixel}$, which results in an error of less than 0.3 mm. In this article, the target value for a crack edge pixel is 1, while the rest is 0.

\subsection{Verification of Ground-Truth Labels}

The above-described procedure may falsely label some pixels as crack edges under certain circumstances. For example, suppose an I-FIT specimen has irregularities and holes on the surface that cannot be painted or create shadows. In that case, the error in DIC measurements will increase compared to experiments on flat surfaces. This may lead to falsely labeled crack edge pixels. To make the ground-truth crack edges as accurate as possible, every image went through two rounds of inspection:
\begin{enumerate}
    \item Automated inspection: for a sequence of deformed images, if pixel $A$ was labeled as a crack edge in frame $F_{t_0}$, but not in the following frames $F_{t_{1:n}}$, the label would be corrected. In contrast, if pixel $A$ was not labeled as crack edge in frame $F_{t_n}$, but was labeled in previous $F_{t_{0:n-1}}$ and following frames $F_{t_{n+1:2n}}$, the label would be corrected;
    \item Manual inspection: the ground-truths were visually inspected and verified.
\end{enumerate}

\subsection{Image Library Summary}
\begin{figure*}[!t]
\centering
{\includegraphics[width=6.0in]{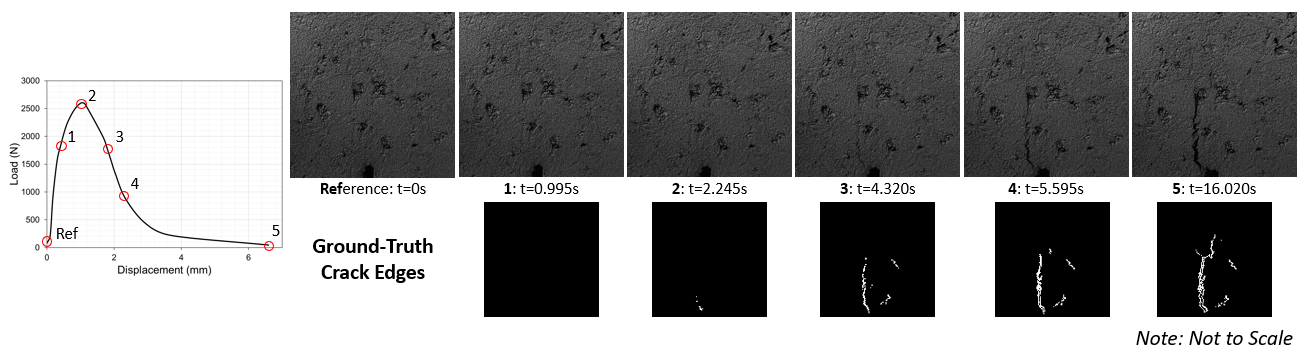}}
\caption{Representative image pairs with ground-truth labels from the image library.}
\label{fig_img_lib}
\end{figure*}

An image library, made of pairs of images with and crack edge labels, was developed for supervised learning. It consisted of 2,560 frame pairs. The image library represented the diversified cracking behavior of AC. Images were collected in the past eight years by four different operators. The original images collected by the Gazelle and the Prosilica camera have resolutions of $6576 \times 4384$ and $2048 \times 2048$, respectively. The original image was downsized to $1024 \times 1024$ by min-pooling and cropping to balance computational overhead and accuracy. This resulted in a spatial resolution of approximately 0.05 $\textnormal{mm/pixel}$. Because the correlation point spacing in DIC analysis was typically larger than 10 pixels, the labeled ground truth crack edges were discontinuous. To provide more accurate ground-truth labels and reduce the computational cost, the ground-truth crack edge map was downsized from the original resolution to $128 \times 128$. This resulted in a spatial resolution of approximately 0.4 $\textnormal{mm/pixel}$, which is adequate in this task, where the measurement area is larger than $50 \times 50 \textnormal{mm}$. Figure \ref{fig_img_lib} illustrates a sequence of images together with their ground-truth crack edge maps. They were collected while conducting an I-FIT test on a typical AC mix (an Illinois N90 mix) at 25 mm/min and 25\textdegree{}C. 

To evaluate the sufficiency of the developed image library, Table \ref{table_img_lib} compared it with datasets used in previous studies for relative tasks. The size of the developed image library is comparable to existing datasets. It is worth noting that StrainNet uses synthetic images, which have the advantage of developing large datasets quickly and inexpensively. However, real images were used in this paper because of the complexity of crack shapes in AC testing.

\begin{table}
\renewcommand{\arraystretch}{1.5}
\tbl{Comparison with datasets used in previous studies for relative tasks.}
{\begin{tabular}{l|l|l|l}
\hline
\hline
\rule{0pt}{10pt}
\textbf{Dataset} & \textbf{Number of frames} & \textbf{Resolution} & \textbf{Type}  \\ \hline
\multicolumn{4}{l}{\emph{Task: Pavement crack detection}} \rule{0pt}{10pt}\\ \hline
CRACK500 \citep{yang2019feature} & 500 & 2000$\times$1500 & Real \\ 
\hline
GAPs384 \citep{eisenbach2017get} & 1,969  & 1920$\times$1080 & Real     \\ 
\hline
CrackNet \citep{zhang2017automated}  & 2,000  & 1024$\times$512 & Real   \\ 
\hline
CrackNet-V \citep{fei2019pixel} & 3,083 & 1024$\times$512 & Real    \\ 
\hline
\multicolumn{4}{l}{\emph{Task: Optical flow estimation}}
\rule{0pt}{10pt}\\ \hline
KITTI2015 \citep{menze2015object} & 800 pairs & 1242$\times$375 & Real \\ 
\hline
Sintel \citep{butler2012naturalistic} & 1064 pairs & 960$\times$540 & Real \\ 
\hline
\multicolumn{4}{l}{\emph{Task: DIC with deep learning}} \rule{0pt}{10pt} \\ \hline
StrainNet \citep{boukhtache2021deep} & 363 reference frames & 256$\times$256 & Synthetic \\ 
\hline
\hline
Proposed  & 2,560 pairs & 1024$\times$1024 & Real \\ 
\hline
\end{tabular}}
\label{table_img_lib}
\end{table}

\section{Network Architecture}
Training a deep CNN from scratch requires a large dataset and significant computational power, which is impossible in most situations. In practice, pre-trained networks could be used as initialization or feature extractors for the task of interest. Because surface cracks are defined as displacement field discontinuities, crack propagation measurement in AC fracture testing can be accomplished by stacking edge detection layers on pre-trained networks for optical flow estimation. 

This section provides a brief overview of existing networks on optical flow estimation. The proposed network architecture is discussed in detail.

\subsection{CNN-Based Methods for Optical Flow Estimation}
Optical flow is the pattern of apparent motion of objects due to the relative motion between an observer and a visual scene \citep{warren2013electronic}. Traditional energy-minimization-based approaches involve computationally expensive optimization, making them unsuitable for large-scale real-time applications such as crack propagation measurement in AC testing, where hundreds of images need to be processed. 

\begin{figure}[!t]
    \centering
    \includegraphics[width=6in]{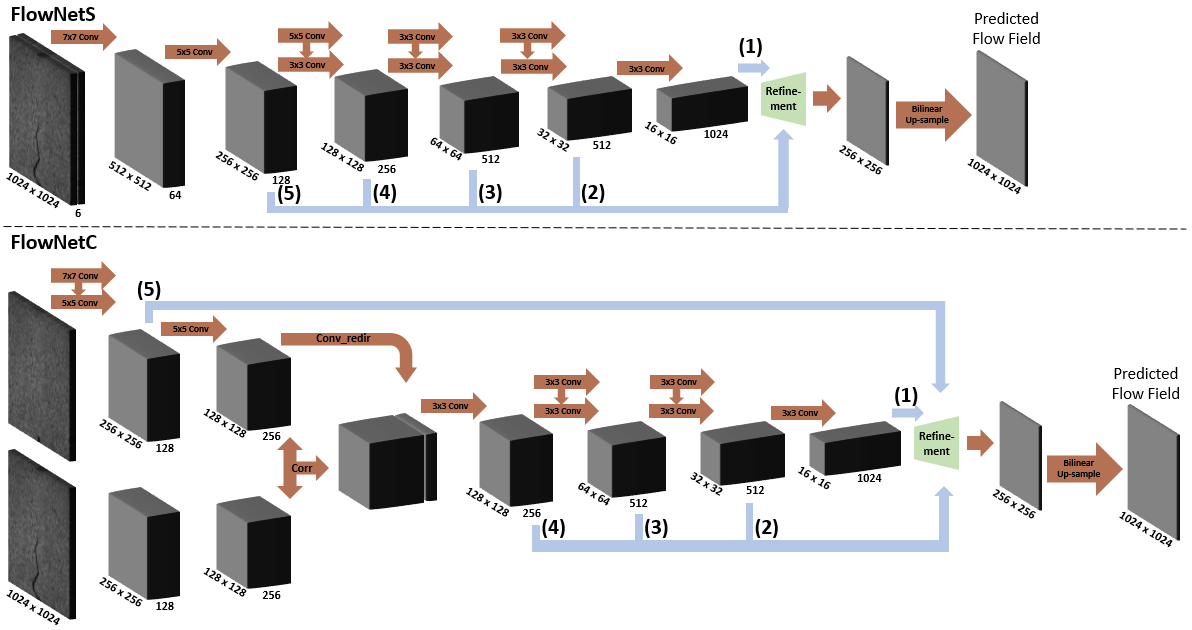}
    \caption{FlowNetS and FlowNetC \cite{dosovitskiy2015flownet}.}
    \label{fig_csnetwork}
\end{figure}

Another promising approach is the fast and end-to-end trainable CNN framework. Dosovitskiy \emph{et al.} proposed two CNNs: FlowNetS and FlowNetC, to learn optical flow from a synthetic dataset \citep{dosovitskiy2015flownet}. As shown in Figure \ref{fig_csnetwork}, FlowNetS stacks the reference and deformed images together and feeds them through a rather generic network to extract optical flow. FlowNetC creates separate processing streams for the reference and deformed images to generate two feature maps. Then, it resembles them with a correlation layer that performs multiplicative patch comparisons. However, FlowNetS and FlowNetC have problems with small displacements and noisy artifacts in estimated optical flow fields. To improve the performance, Ilg \emph{et al.} developed FlowNet2.0 by stacking multiple FlowNetS and FlowNetC \citep{ilg2017flownet}. It reduces the estimation error by more than 50\% compared to FlowNet and has been proven to be efficient in many other applications such as motion segmentation and action recognition. FlowNet2.0 outperforms other state-of-art networks, such as SpyNet, RecSpyNet, and LiteFlowNet \citep{ranjan2017optical,hu2018recurrent,hui2018liteflownet} in terms of accuracy.

The above studies demonstrate that CNNs are powerful in estimating optical flow. This inspired the development of similar networks to solve analogous problems in other fields. For example, Boukhtache \emph{et al.} developed StrainNet to retrieve displacement and strain fields from pairs of reference and deformed images. It uses FlowNetS as the backbone and achieves comparable accuracy as DIC with a significant improvement in computing time \citep{boukhtache2021deep}.

\begin{figure}[!t]
    \centering
    \includegraphics[width=6in]{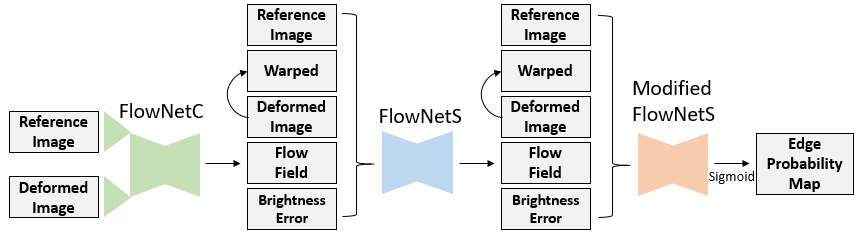}
    \caption{Proposed network architectures.}
    \label{fig_network}
\end{figure}

\subsection{FlowNetS and FlowNetC}
Both FlowNetS and FlowNetC consist of a contracting and an expanding part. Although the two networks adopt different approaches in contracting, they share the same expanding part. The two parts of the network are discussed in detail below.

\subsubsection{Contracting}
FlowNetS concatenates the reference and deformed images together as input and lets the network learn how to process the image pair to extract motion information. In contrast, FlowNetC creates separate processing streams for the reference and deformed images to generate two feature maps. Then, it resembles them with a correlation layer that performs multiplicative patch comparisons. Given two feature maps $\textbf{f}_1$, $\textbf{f}_2$, with dimension $c (\textnormal{number of channels}) \times w (\textnormal{width}) \times h (\textnormal{height})$, the correlation layer compares patches in $\textbf{f}_1$ and $\textbf{f}_2$ as below: 

\begin{equation}
        c(\textbf{x}_1, \textbf{x}_2) = \sum_{\textbf{o} \in [-k,k] \times [-k,k]} \langle \textbf{f}_1(\textbf{x}_1+\textbf{o}), \textbf{f}_2(\textbf{x}_2+\textbf{o}) \rangle
    \label{eqn_corr}
\end{equation}

$\textbf{x}_1$ and $\textbf{x}_2$ denote the center square patch of size $K \coloneqq 2k+1$ in $\textbf{f}_1$ and $\textbf{f}_2$, respectively. Because the correlation operation is fundamentally equivalent to convolving data with other data, it has no trainable weights. To reduce computation cost, the maximum displacement is constrained to $d$, which means that for each location $\textbf{x}_1$, correlations $c(\textbf{x}_1, \textbf{x}_2)$ are only computed in a neighborhood of size $D \coloneqq 2d+1$. The output size of the correlation layer is $D^2 \times w \times h$.

\subsubsection{Expanding}
\begin{figure}[!t]
    \centering
    \includegraphics[width=5.0in]{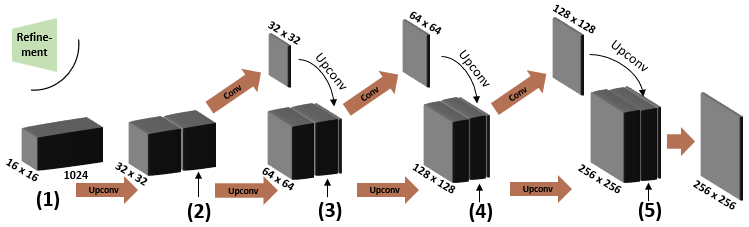}
    \caption{Refinement part.}
    \label{fig_refine}
\end{figure}

As shown in Figure \ref{fig_refine}, to refine the coarse pooled representation and obtain a dense flow field, fractionally-strided convolution is applied to feature maps first. Then, they are concatenated with corresponding feature maps from the contracting part. This operation preserves high-level features as well as retains fine local features. The process is repeated four times, with each step doubling the resolution. A bi-linear up-sampling with a factor of four is performed at the end to obtain the original image resolution.

\subsection{Proposed Network Architecture}
\begin{figure}[!t]
    \centering
    \includegraphics[width=3.45in]{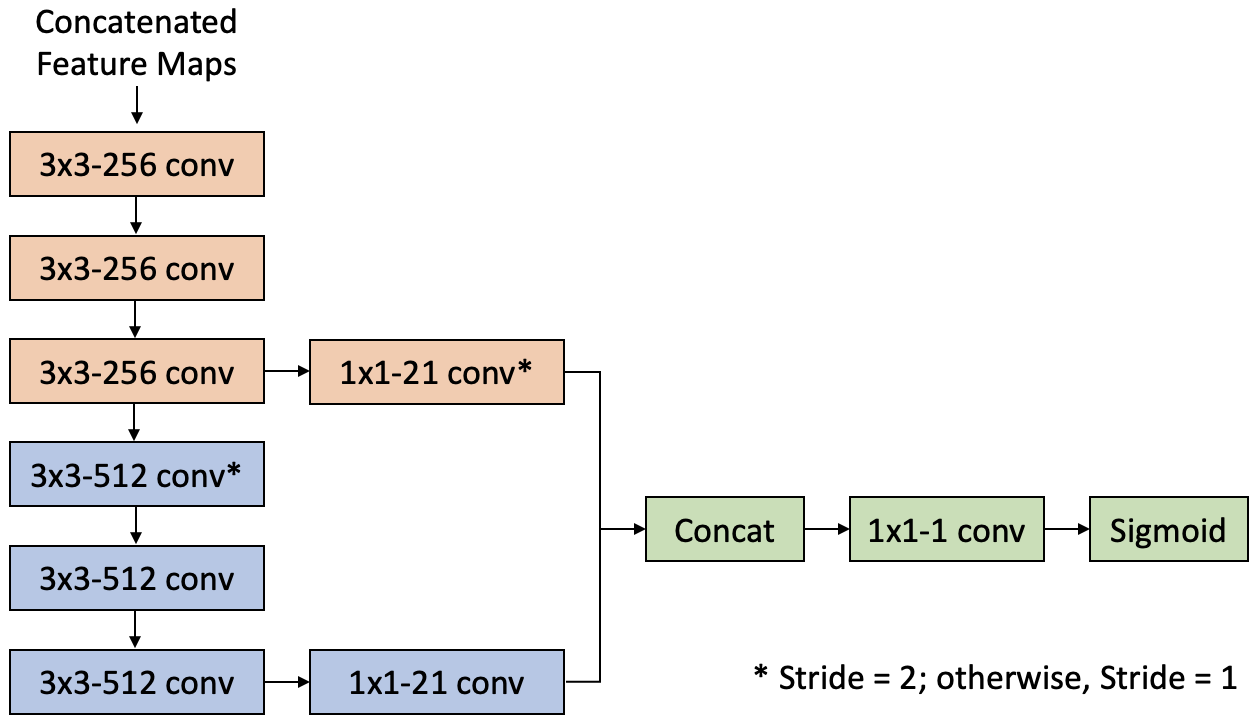}
    \caption{Edge detection layers.}
    \label{fig_edge}
\end{figure}

As shown in Figure \ref{fig_network}, the proposed network combines one FlowNetC and two FlowNetS. First, reference ($I_r$) and deformed ($I_d$) images are fed into FlowNetC to generate an estimated flow field ($w_1 = (u_1,v_1)^ \top$) at the original image resolution. Second, subsequent FlowNetS gets reference image, deformed image, estimated flow field $w_1$, warped deformed image ($\tilde I_{d,1} (x,y) = I_d(x+u_1,y+v_1)$), and brightness error field ($e_1 = ||\tilde I_{d,1} - I_r||$); and outputs an estimated flow field ($w_2 = (u_2,v_2)^ \top$) at the original image resolution. The concatenated input allows FlowNetS to assess the previous error more easily and compute an incremental update. Third, a modified FlowNetS receives reference image, deformed image, estimated flow field $w_2$, warped deformed image ($\tilde I_{d,2} (x,y) = I_d(x+u_2,y+v_2)$), and brightness error field ($e_2 = ||\tilde I_{d,2} - I_r||$); and outputs an edge probability map for which the resolution is eight times smaller than the original image. The modified FlowNetS differs from FlowNetS in the expanding part: 
\begin{itemize}
    \item The last $3 \times 3$ convolution layer is cut. Inspired by Richer Convolutional Features for edge detection \citep{liu2017richer}, layers shown in Figure \ref{fig_edge} are added.
    \item Sigmoid units are connected to the final layer to generate an edge probability map.
\end{itemize}

\section{Training}
\subsection{Class-Balanced Loss Function}
Because the distribution of crack-edge/non-crack-edge pixels greatly varies and is heavily biased: more than 90\% of the ground truth pixels are non-crack-edge. A cost-sensitive loss function must be considered to balance the loss between positive (crack-edge)/negative (non-crack-edge) classes. Specifically, the following (Equation \ref{eqn_loss}) class-balanced cross-entropy loss function was used:

\begin{equation}
    \begin{aligned}
        L(W) &= \alpha \sum_{j \in Y_-}\log(1- \Pr (X_j; W)) \\ &+ \beta \sum_{j \in Y_+}\log \Pr (X_j; W)
    \end{aligned}
    \label{eqn_loss}
\end{equation}

in which

\begin{equation}
    \begin{gathered}
        \alpha = \gamma + \frac{|Y_+|}{|Y_+|+|Y_-|} \\
        \beta = \lambda \cdot \frac{|Y_-|}{|Y_+|+|Y_-|}
    \end{gathered}
    \label{eqn_alpha}
\end{equation}

$Y_+$ and $Y_-$ denote the positive and negative sample sets, respectively. The hyperparameters $\gamma$ and $\lambda$ are used to balance positive and negative samples. $X_j$ represents the activation value at each pixel $j$, $\Pr (X)$ is the standard sigmoid function, and $W$ denotes all parameters in the network. 

\subsection{Data Augmentation}
Data augmentation is an often-used strategy to improve model generalization \citep{krizhevsky2012imagenet}. The augmentations used in this study include geometric transformation: horizontal flip, as well as changes in brightness, contrast, saturation, and hue. It is worth mentioning that the same transformations were applied to both reference and deformed images. The augmentation was performed online during network training. 

The brightness, contrast, and saturation factors are sampled uniformly from $[0.95, 1.05]$; the hue factor is chosen from $[-0.05,0.05]$. 

\subsection{Training Strategy}
The following strategy was used in training the network:
\begin{enumerate}
    \item The AdamW was chosen as the optimization method because it showed faster convergence than standard stochastic gradient descent with momentum in this task \citep{loshchilov2017decoupled}. The recommended parameters: $\beta_1 = 0.9$ and $\beta_2 = 0.999$ were used, and the weight decay coefficient was set as $1 e^{-4}$. 
    \item Fairly small mini-batches of six-image pairs were used.
    \item The training started with a learning rate of $5e^{-5}$, and it was divided by 2 every 5 epochs. The network was trained for 40 epochs.
    \item To monitor over-fitting during training, the dataset was randomly split into 2,248 training and 312 validation pairs.
\end{enumerate}

\subsection{Evaluation Metrics}
Because of the similarity with edge detection, it is intuitive to directly leverage its evaluation criteria for this task. Given a crack edge probability map, a threshold is needed to generate the crack edge map. Two commonly used strategies are optimal dataset scale (ODS) and optimal image scale (OIS). The former uses a fixed threshold for all images in the dataset, while the latter employs an optimal threshold for each image \citep{xie2015holistically,liu2017richer}. This paper used the F-1 ($\frac{2 \cdot Precision \cdot Recall}{Precision + Recall}$) of both ODS and OIS to assess the network's performance. They were calculated using Equation \ref{eqn_ods_ois} \citep{yang2019feature}. It is worth noting that, unlike previous studies, zero tolerance was allowed for correct matches between ground truth and prediction \citep{xie2015holistically, liu2017richer, yang2019feature}.

\begin{equation}
    \begin{gathered}
        \textnormal{ODS F} = \textnormal{max}\{\frac{2P_t \cdot R_t}{P_t + R_t} : t = 0.01,0.02, \dots, 0.99\} \\
        \textnormal{OIS F} = \frac{1}{N_{i}} \sum_i^{N_{i}} \textnormal{max}\{\frac{2P_t^i \cdot R_t^i}{P_t^i + R_t^i} : t = 0.01,0.02, \dots, 0.99\}
    \end{gathered}
    \label{eqn_ods_ois}
\end{equation}

The $t$ denotes the threshold, $i$ refers to the index of an image, and $N_{i}$ is the total number of images. $P_t$ and $R_t$ represent precision and recall for the chosen threshold $t$, respectively. Precision refers to the proportion of identified crack edge pixels that were correct, while Recall represents the fraction of crack edge pixels identified correctly. It is challenging to achieve high Precision and high Recall simultaneously because they often conflict with each other. A high F-1 can only be achieved when both Precision and Recall are high.

\subsection{Training Result}
\begin{figure}[!t]
    \centering
    \includegraphics[width=3.45in]{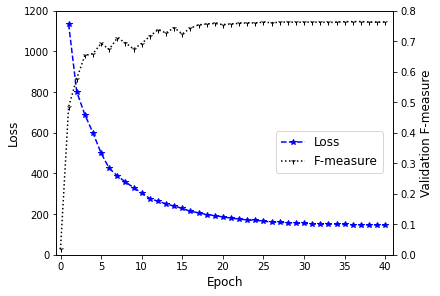}
    \caption{Training progress.}
    \label{fig_train}
\end{figure}

\begin{figure}[!t]
    \centering
    \includegraphics[width=3.in]{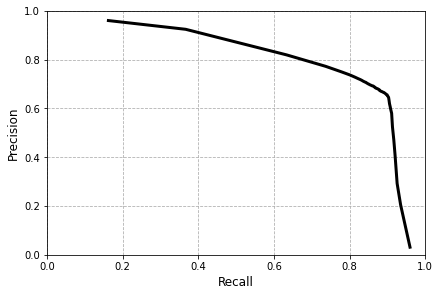}
    \caption{Precision-recall curve evaluated on the validation dataset.}
    \label{fig_pr}
\end{figure}

The training took 13 hours on an \emph{NVIDIA TESLA V100} GPU. Figure \ref{fig_train} shows the class-balanced cross-entropy loss decay curve and the validation F-1 curve. To compute the F-1 during training, a fixed threshold of 0.5 was used to generate edge maps from edge probability maps instead of using ODS or OIS strategies. The highest F-1 was observed at the $37^{th}$ epoch, and the corresponding model was considered optimal. Figure \ref{fig_pr} shows the precision-recall curve of the final model. The trained model achieved $\textnormal{ODS}=0.769$ and $\textnormal{OIS}=0.772$ on the validation dataset.

\section{Testing and Evaluation}
\subsection{Testing Result}
\begin{figure}[!t]
    \centering
    \includegraphics[width=3in]{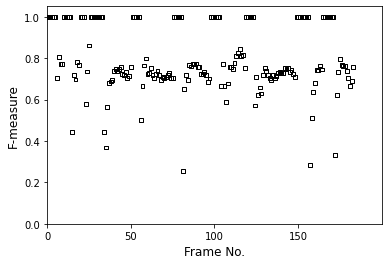}
    \caption{F-1s on each frame pair.}
    \label{fig_f1}
\end{figure}

\begin{figure}[!t]
    \centering
    \includegraphics[width=4in]{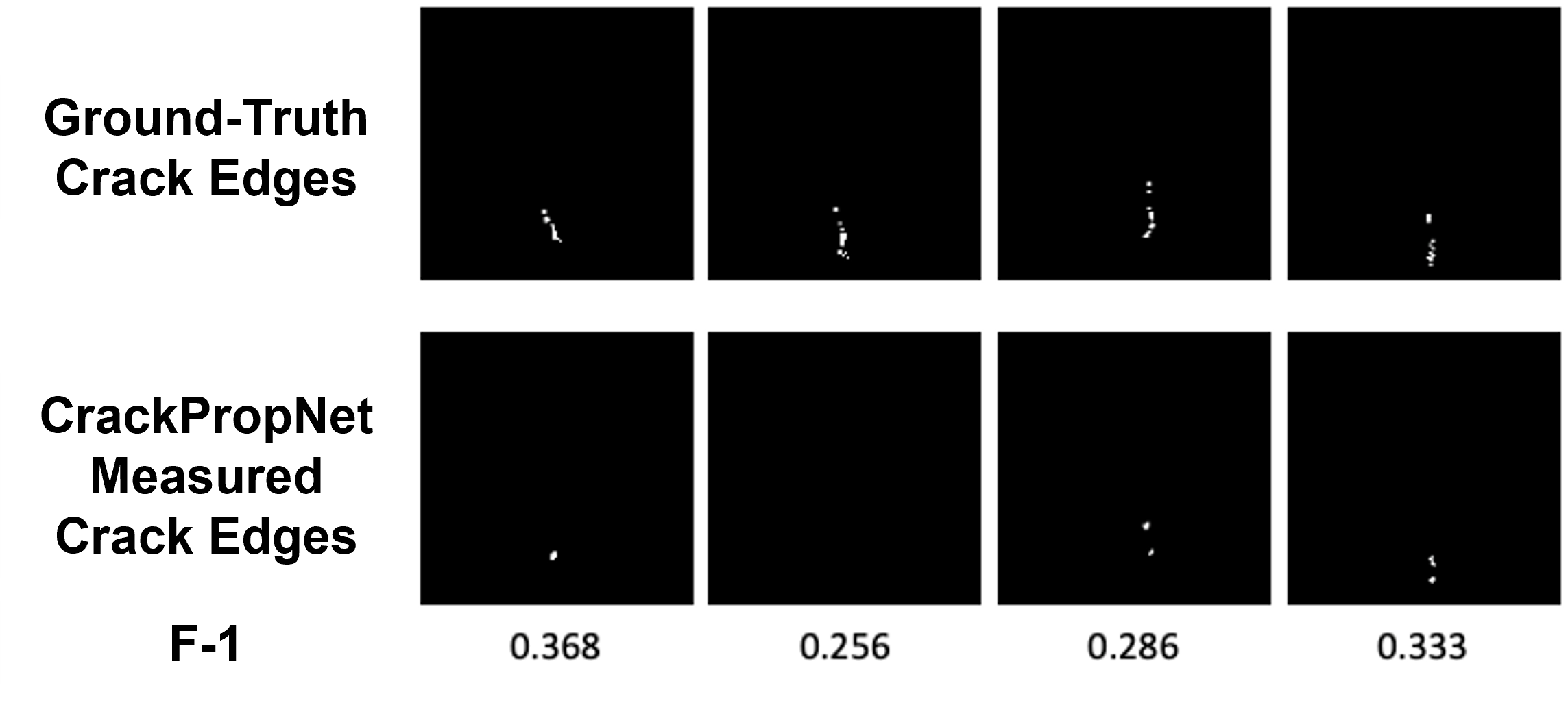}
    \caption{CrackPropNet-measured cracks with low F-1s.}
    \label{fig_bad_pred}
\end{figure}

A testing dataset consisting of 188 frame pairs was developed to validate the trained model further. The images were collected using the Gazelle and the Prosilica cameras while conducting I-FIT tests. The same ground-truth crack edges labeling procedure was followed. CrackPropNet provided running speeds of 6fps and 26fps on a \emph{NVIDIA TESLA P100} and a \emph{TESLA V100} GPU, respectively. The trained model achieved $\textnormal{ODS}=0.755$ and $\textnormal{OIS}=0.781$ on the testing dataset. The validation and testing images' performance were similar, suggesting that the over-fitting problem was avoided. Figure \ref{fig_f1} shows F-1s on each frame pair. The trained model generally performed well in most frame pairs. Less than 2\% of edge map predictions had F-1s smaller than 0.4. Moreover, the model performed exceptionally well in differentiating the frames with no cracks from those with cracks, which indicates its robustness in capturing crack initiation. Figure \ref{fig_bad_pred} shows edge map predictions with F-1s lower than 0.4. It could be noticed that all of them happened in the early stage of crack development. 

\begin{figure*}[!t]
\centering
{\includegraphics[width=6.0in]{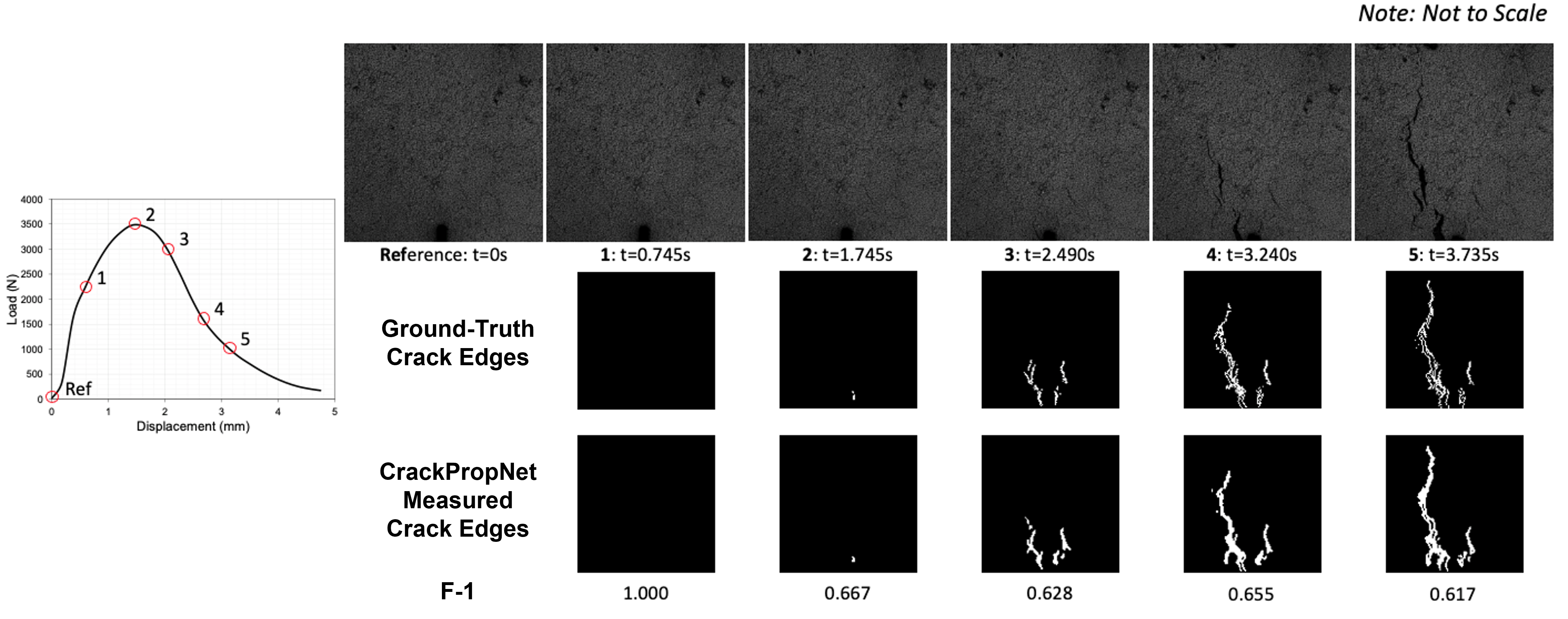}}
\caption{Examples of CrackPropNet-measured crack propagation.}
\label{fig_vis_pred}
\end{figure*}

Figure \ref{fig_vis_pred} provides a visualization of a sequence of CrackPropNet-measured crack edges. They were intentionally selected to be shown here because of their lower-than-average F-1s. The final model produced high-quality crack edges. The lower-than-average F-1 was mainly due to the nature of edge detection, where predicted edges are expected to be coarser than ground-truths \citep{liu2017richer, xie2015holistically}. 

\subsection{Noise Robustness Evaluation}
Some noise is always present in digital images, especially for those taken by  nonindustrial low-cost cameras. Because the training images were collected using high-performance hardware, it is critical to evaluate the noise robustness of the trained model. 

Random Gaussian noise was injected to frame pairs in the testing dataset based on the Gaussian noise model:

\begin{equation}
        P(g) = \sqrt{\frac{1}{2\pi\sigma^2}}e^{-\frac{(g-\mu)^2}{2\sigma^2}}
    \label{eqn_gaussian}
\end{equation}

\begin{figure}[!t]
    \centering
    \includegraphics[width=2.5in]{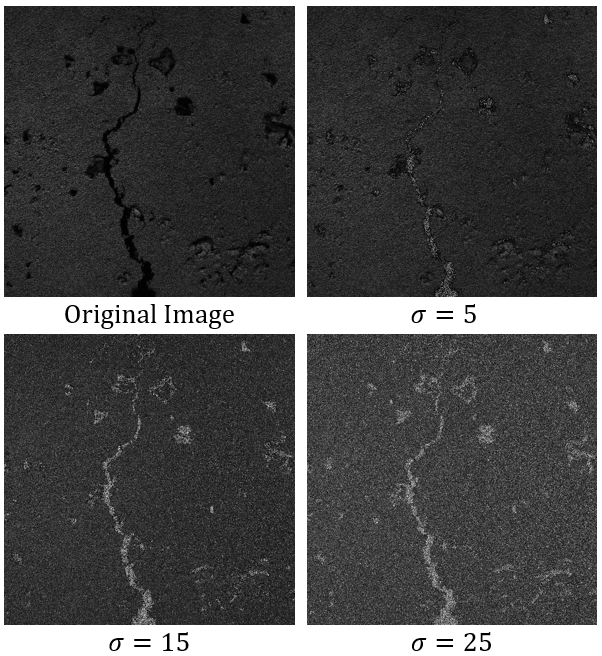}
    \caption{Images with different levels of random Gaussian noises were injected.}
    \label{fig_noise_img}
\end{figure}

$\mu$ and $\sigma$ denote mean and standard deviation, respectively. $g$ refers to the gray value. Three $\sigma$ (5,15,25) values were used to simulate various degrees of noise: low, medium, and high. Figure \ref{fig_noise_img} shows images with different levels of random Gaussian noises injected. 

Table \ref{table_noise} shows the model performance on noise-injected images in the testing dataset. As would be expected, the measurement accuracy decreases as $\sigma$ increases. The trained model performs well on images with low to medium noise levels. 

\begin{table}
\renewcommand{\arraystretch}{1.5}
\tbl{Model performance on testing images with noise injected.}
{\begin{tabular}{l|l|l}
\hline
\hline
$\sigma$ & \textbf{ODS} & \textbf{OIS} \\ 
\hline
5 & 0.6643 & 0.7720 \\ 
\hline
15 & 0.5918 & 0.6586 \\ 
\hline
25 & 0.5292 & 0.5915 \\ 
\hline
\end{tabular}}
\label{table_noise}
\end{table}

\subsection{IDEAL-CT}
\begin{figure}[!t]
    \centering
    \includegraphics[width=2.in]{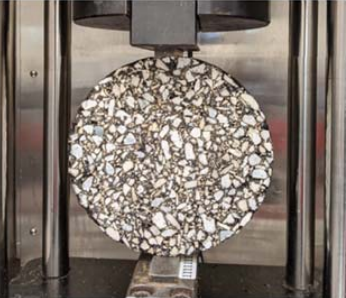}
    \caption{IDEAL-CT setup.}
    \label{fig_ideal}
\end{figure}

\begin{figure*}[!t]
\centering
{\includegraphics[width=6.0in]{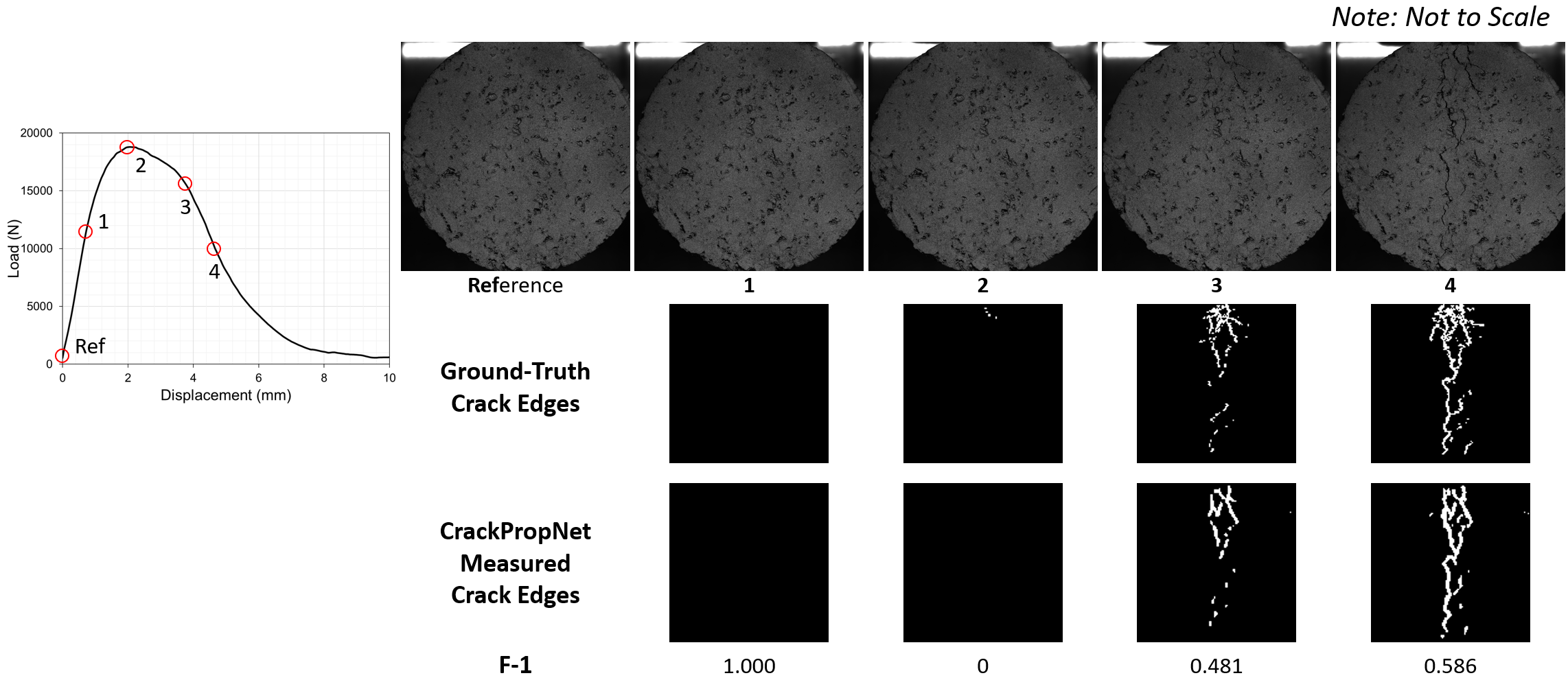}}
\caption{Representative IDEAL-CT image pairs with ground-truth and CrackPropNet-measured crack edges.}
\label{fig_ideal_vis}
\end{figure*}

To evaluate the generalization of the trained model, a small dataset consisting of 91 frame pairs was developed. The images were collected using the Prosilica ($6576 \times 4384$ pixels, 4 fps) while conducting the indirect tensile cracking test (IDEAL-CT), as shown in Figure \ref{fig_ideal}. The spatial resolution was about 35 $\mu\textnormal{m/pixel}$. The test was performed at 25\textdegree{}C and 50 mm/min LLD on a cylindrical specimen of 62 mm in thickness and 150 mm in diameter. The IDEAL-CT is fundamentally different from the I-FIT test. The former is a strength test requiring no notch, while the latter is a fracture test with a pre-crack (notch). Most I-FIT specimens have a single, well-defined crack path, unlike the IDEAL-CT specimens with multiple crack paths \citep{al2021cracking}. Moreover, as shown in Figure \ref{fig_ideal_vis}, the images contained blurred backgrounds, which poses a new challenge to the trained model.

Figure \ref{fig_ideal_vis} illustrates a sequence of images with their ground-truth crack edge maps. They were collected while conducting an IDEAL-CT test on an AC mix with high asphalt binder replacement (20\%) at 50 mm/min and 25\textdegree{}C.

\begin{figure}[!t]
    \centering
    \includegraphics[width=3.in]{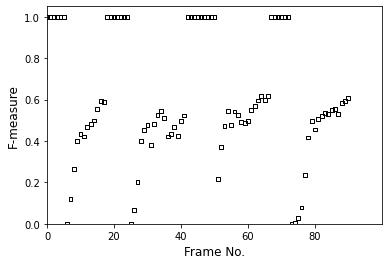}
    \caption{F-1s on each frame pair of the IDEAL-CT dataset.}
    \label{fig_idealct_ind_f1}
\end{figure}

The trained model achieved $\textnormal{ODS}=0.588$ and $\textnormal{OIS}=0.605$ on the evaluation dataset. Figure \ref{fig_ideal_vis} provides a visualization of a sequence of CrackPropNet-measured  crack edges. Overall, the trained model showed promising accuracy on a dataset that is fundamentally different from the training dataset.  The measurement accuracy increased as the crack propagated downwards. The trained model was able to measure fine details of mature cracks, as shown in Figure \ref{fig_ideal_vis}, frames 3 and 4. Figure \ref{fig_idealct_ind_f1} shows F-1s on each frame pair. The relatively low overall F-1 was mainly due to the poor performance on small cracks in the early stage of development, where multiple crack paths were presented in an IDEAL-CT strength test specimen. 

The promising accuracy of the CrackPropNet in the case of IDEAL-CT indicated the model was well-trained to locate displacement field discontinuity, which is the definition of crack. As would be expected, the CrackPropNet could provide a relatively accurate measurement of crack propagation in other AC cracking tests regardless of the cracking mechanisms.

\subsection{Application: Crack-Propagation Speed}
As a crack measurement technique, the CrackPropNet can detect complex crack patterns accurately and efficiently in AC cracking tests. The trained model was applied to calculate crack-propagation speed in a fracture test as a case study to demonstrate its usefulness. 

Crack-propagation speed is one of the main AC cracking characteristic factors. A large crack-propagation speed after initiation indicates the mix is brittle and prone to cracking. Most state-of-art AC cracking potential prediction indices rely on an approximate crack-propagation speed. For example, according to AASHTO T393, the flexibility index (FI) from the Illinois-flexibility index test (I-FIT) uses the post-peak inflection-point slope from the Load-LLD curve to proxy the crack-propagation speed. The speed was assumed constant \citep{al2015testing}. With the help of the trained model, the true crack-propagation speed could be easily derived and used to calculate cracking indices.

This case study included two plant-produced AC mixes, and their design details are summarized in Table \ref{table_mix}. Raw images were collected while conducting the I-FIT test at 50 mm/min and 25\textdegree{}C. Four replicates were used for each mix. It was expected that mix two would have a much higher crack propagation speed than mix one because: 
\begin{itemize}
    \item Mix two had significantly lower asphalt content than mix one.
    \item Mix two used recycled materials, while mix one did not. 
\end{itemize}

\begin{table}
\renewcommand{\arraystretch}{1.5}
\tbl{Mix Design Details for AC Mixes Used in This Study.}
{\begin{tabular}{l|l|l}
\hline
\hline
Property & Mix 1 & Mix 2 \\ 
\hline
Type & SMA & Dense-Graded \\ 
\hline
Binder Grade & PG 70-22 & PG 64-22 \\ 
\hline
Asphalt Content (\%) & 7.3 & 4.98 \\
\hline
NMAS(mm) & 4.75 & 12.5 \\ 
\hline
ABR (\%) & 0 & 20 \\ 
\hline
VMA (\%) & 18.5 & 14.6 \\ 
\hline
\end{tabular}}
\label{table_mix}
\end{table}

\begin{figure}[!t]
    \centering
    \includegraphics[width=4.5in]{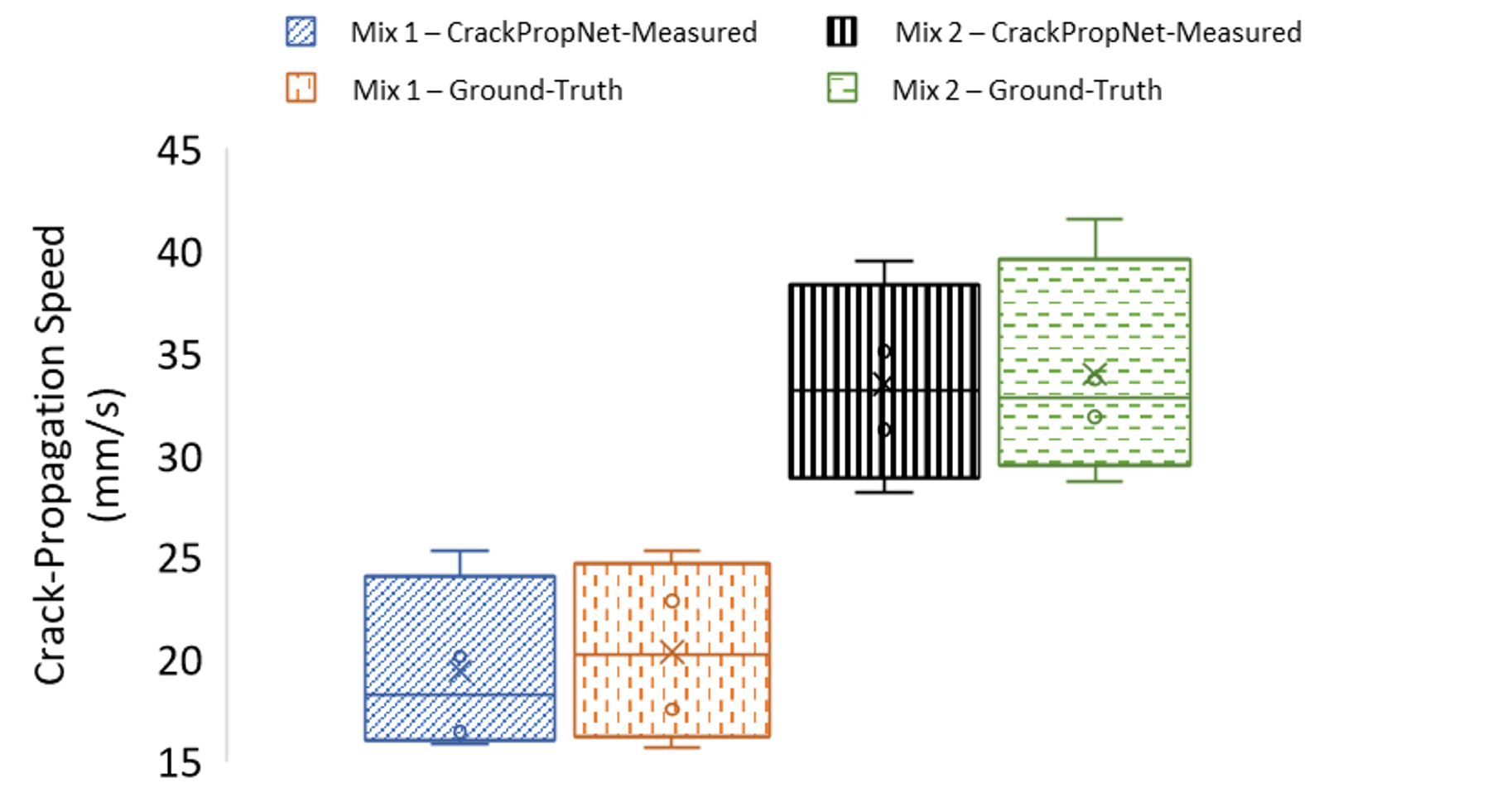}
    \caption{CrackPropNet-measured and ground-truth mean crack-propagation speed of mix one and two.}
    \label{fig_crack_speed}
\end{figure}

Figure \ref{fig_crack_speed} shows the CrackPropNet-measured and ground-truth mean crack-propagation speed of AC mixes one and two. The speed was calculated by tracking the crack front and averaged along the crack path. As would be expected, the mean crack-propagation speed measured on mix two specimens was 72\% faster than that on mix one, indicating that mix two is more prone to cracking than mix one. The CrackPropNet-measured mean crack-propagation speed was similar to the ground-truth crack-propagation speed. The trained model achieved a mean absolute error of 1.08 mm/s on the tested specimens. Moreover, the CrackPropNet captured the AC material-inherent crack propagation speed variability.

\section{Discussion}
As summarized in Table \ref{table_comparison}, the CrackPropNet offers an accurate, flexible, efficient, and low-cost solution for crack propagation measurement in AC cracking tests. To measure crack propagation on AC specimen surfaces, CrackPropNet only needs a series of images collected by a low-cost camera while conducting cracking tests and a computer with GPU for post-processing.

\begin{table}
\renewcommand{\arraystretch}{1.5}
\tbl{Comparison between the CrackPropNet and other crack measurement techniques.}
{\begin{tabular}{l|l|l|l|l}
\hline
\hline
Technique & Accuracy & Flexibility & Efficiency & Cost \\ 
\hline
Contact Tools (e.g., LVDT, clip gauge)& High & Low & Medium & $\sim$\$1,000 \\ 
\hline
Low-Level Computer Vision (e.g., thresholding) & Low & Medium & Medium & $\sim$\$500 \\ 
\hline
Digital Image Correlation \citep{zhu2022} & High & High & Low & $\sim$ \$10,000 \\
\hline
CrackPropNet & High\textsuperscript{a} & High & High & $\sim$ \$1,500 \\ 
\hline
\end{tabular}}
\tabnote{\textsuperscript{a}Comparable to digital image correlation.}
\label{table_comparison}
\end{table}

Although the CrackPropNet was trained on an image database of I-FIT tests, its promising measurement accuracy in the case of IDEAL-CT suggested that it could provide a relatively accurate measurement of crack propagation in other AC cracking tests. This is because the architecture of the CrackPropNet was designed to learn to locate displacement field discontinuities (i.e., cracks) regardless of the cracking mechanism.

CrackPropNet has many applications. Examples are listed below:
\begin{itemize}
    \item Compare AC mixes' cracking potential. In a C* fracture test, a video of the specimen surface is recorded during the test, and crack propagation is measured via visual recognition, which is subjective and time-consuming \citep{stempihar2013development}. Instead, one can use CrackPropNet to measure crack propagation. An AC mix with a faster crack propagation speed indicates that it is more prone to cracking.
    \item Validate test protocols. To validate the testing protocol of the single-edge notched beam (SENB) test, wire crack detection gauges were glued to the specimen surface to monitor crack propagation \citep{wagoner2005development}. The gauges only provided localized information. Instead, CrackPropNet can be used for full-field crack propagation measurement.
    \item Derive new or calibrate existing cracking indices. The FI, the primary outcome of the I-FIT, introduced the post-peak inflection-point slope to proxy the crack-propagation speed after initiation. With the development of CrackPropNet, the actual crack propagation can be efficiently measured and used to compute cracking indices. Similarly, it can be used to calibrate existing indices.
\end{itemize}

\section{Conclusions}
This article proposes an efficient deep neural network, namely CrackPropNet, to measure crack propagation on AC specimens during testing. The proposed approach provides accuracy, flexibility, efficiency, and cost-effectiveness compared to other techniques, including contact measurements, low-level computer vision, and DIC.

CrackPropNet involves learning to locate displacement field discontinuities (i.e., cracks) by matching features at various locations in the reference and deformed images. The input of CrackPropNet includes a reference and a deformed image, and a crack edge probability map is generated as the output. This was accomplished by stacking edge detection layers on a pre-trained optical flow estimation network consisting of a FlowNetC and two FlowNetS.

An image library was developed for supervised learning. It represents the diversified AC cracking behavior. CrackPropNet could provide running speeds of 6fps and 26fps on a \emph{NVIDIA TESLA P100} and a \emph{TESLA V100} GPU, respectively. Besides, it achieved promising measurement accuracy with an ODS F-1 of 0.755 and an OIS F-1 of 0.781 on the testing dataset. 

It was demonstrated in an experiment that low to medium-level Gaussian noises had a limited impact on the measurement accuracy of CrackPropNet. Besides, the model showed promising performance on a fundamentally different dataset, which consisted of images collected while conducting the IDEAL-CT strength test. A case study demonstrated that CrackPropNet could accurately calculate crack-propagation speed, one of the main AC cracking characteristics.

CrackPropNet has many applications, including characterizing the cracking phenomenon, evaluating AC cracking potential, validating test protocols, and verifying theoretical models. The promising performance of CrackPropNet suggests that an optical flow-based deep learning network offers a robust solution in accurately and efficiently measuring crack propagation on AC specimens. 

The followings are recommended for further research:
\begin{itemize}
    \item Images collected during the process of crack propagation possess sequential nature. It is worth investigating if deep recurrent optical flow neural networks could boost the accuracy.
    \item The accuracy and generalization of CrackPropNet are limited by the size of the database, which could be expanded by collecting more images from various AC tests.
    \item Because of AC’s material-inherent variability, it is suggested to monitor both sides of a specimen for the test replicates to obtain a reliable crack propagation measurement.
    \item It is recommended to verify existing cracking indices using CrackPropNet. It would enable contractors and transportation agencies to assess the cracking potential of AC mixes more accurately and efficiently.
\end{itemize}

\section*{Data Availability Statement}
Examples of the image database and pre-trained CrackPropNet are available at \url{https://github.com/zehuiz2/CrackPropNet}.

\section*{Author Contributions}
The authors confirm their contribution to the paper as follows:  study conception and design: Zehui Zhu and Imad L. Al-Qadi; data collection: Zehui Zhu; analysis and interpretation of results: Zehui Zhu and Imad L. Al-Qadi; draft manuscript preparation: Zehui Zhu and Imad L. Al-Qadi. All authors reviewed the results and approved the final version of the manuscript.

\section*{Acknowledgment}

The authors would like to thank Jose Julian Rivera Perez, Berangere Doll, Uthman Mohamed Ali, and Maxwell Barry for their help in preparing test specimens and collecting raw images. The contents of this report reflect the view of the authors, who are responsible for the facts and the accuracy of the data presented herein.

\section*{Disclosure statement}
The authors report there are no competing interests to declare.

\bibliographystyle{apacite}
\bibliography{interactapasample}

\end{document}